\documentclass[10pt,twocolumn,letterpaper]{article}
\pdfoutput=1

\usepackage{iccv}
\usepackage{times}
\usepackage{epsfig}
\usepackage{graphicx}
\usepackage{amsmath}
\usepackage{amssymb}
\usepackage{multirow}
\usepackage{tabularx}
\usepackage[breaklinks=true,bookmarks=false]{hyperref}

\newcommand{\Fig}{Fig.\@\xspace}
\newcommand{\Tab}{Tab.\@\xspace}
\newcommand{\Eq}{Eq.\@\xspace}
\newcommand{\myparagraph}[1]{\vspace{0.5em}\noindent\textbf{#1}}

\newcommand{\nR}{\mathbb{R}}
\newcommand{\nRplus}{\nR_{\geq 0}}
\newcommand{\nProj}{\Pi} 
\newcommand{\nImg}{I}
\newcommand{\nImgDom}{\Omega}
\newcommand{\nSurf}{S}
\newcommand{\nDepth}{\mathbf{d}}
\newcommand{\nNormal}{\mathbf{n}}
\newcommand{\nVertices}{\mathcal{V}}
\newcommand{\nFaces}{\mathcal{F}}
\newcommand{\nVertex}{\mathbf{v}}
\newcommand{\nEdge}{\mathbf{e}}
\newcommand{\nFace}{f}
\newcommand{\nNeigh}{\mathcal{N}}
\newcommand{\nNumLabels}{L}
\newcommand{\nLabel}{l}
\newcommand{\nLabeling}{\mathbf{l}}
\newcommand{\nLabelSet}{\mathcal{L}}
\newcommand{\nLabelImg}{\nImg^\nLabel} 

\iccvfinalcopy 
\def\iccvPaperID{0000}

\ificcvfinal\fi
\begin{document}

\title{Semantically Informed Multiview Surface Refinement}

\author{Maro\v{s} Bl\'aha \hspace{0.25cm} Mathias Rothermel \hspace{0.25cm} Martin R. Oswald \hspace{0.25cm} Torsten Sattler \\ Audrey Richard \hspace{0.25cm} Jan D. Wegner \hspace{0.25cm} Marc Pollefeys \hspace{0.25cm} Konrad Schindler\\
ETH Zurich\\
}

\maketitle

\begin{abstract}
We present a method to jointly refine the geometry and semantic
segmentation of 3D surface meshes.
Our method alternates between updating the shape and the semantic
labels.
In the geometry refinement step, the mesh is deformed with variational energy
minimization, such that it simultaneously maximizes photo-consistency
and the compatibility of the semantic segmentations across a set of
calibrated images. Label-specific shape priors account for
interactions between the geometry and the semantic labels in 3D.
In the semantic segmentation step, the labels on the mesh are updated
with MRF inference, such that they are compatible with the semantic
segmentations in the input images. Also, this step includes prior
assumptions about the surface shape of different semantic classes.
The priors induce a tight coupling, where semantic information
influences the shape update and vice versa.
Specifically, we introduce priors that favor (i) adaptive smoothing,
depending on the class label; (ii) straightness of class boundaries;
and (iii) semantic labels that are consistent with the surface
orientation.
The novel mesh-based reconstruction is evaluated in a series of
experiments with real and synthetic data. We compare both to
state-of-the-art, voxel-based semantic 3D reconstruction, and to
purely geometric mesh refinement, and demonstrate that the proposed
scheme yields improved 3D geometry as well as an improved semantic
segmentation.
\end{abstract}

\section{Introduction}\label{sec:intro}

\begin{figure} 
  \centering
  \includegraphics[width=0.99\columnwidth]{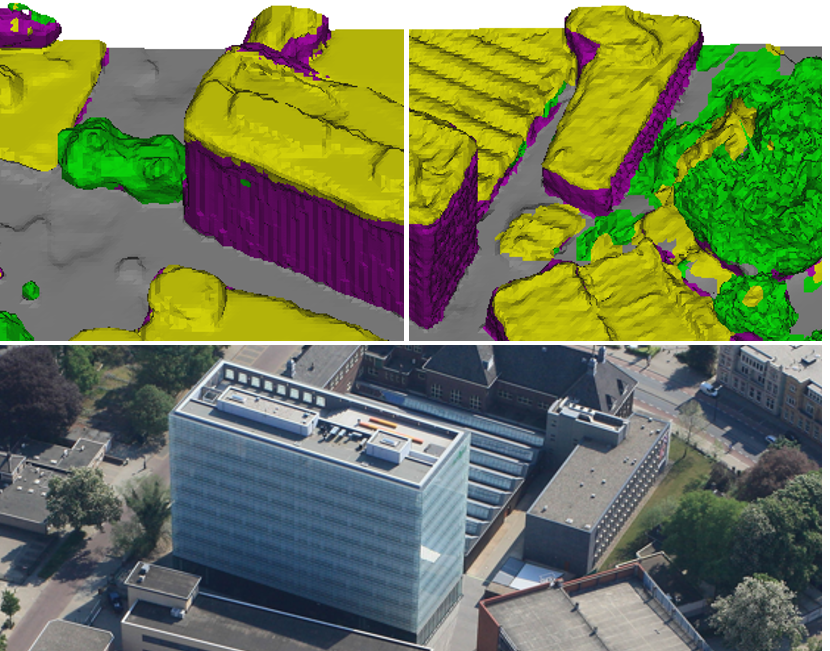}
  \caption{The impact of the proposed method. \textit{Top}: labeled input model and the result after performing geometric and semantic surface refinement (\textit{left-right}). Notice the higher scene fidelity - \eg emerging structures on roofs and facades - and simultaneously an adequate, class-specific regularization in our mesh. \textit{Bottom}: corresponding image.}
  \label{fig:front}
\end{figure}

Extracting 3D scene models from multiple images is one of the core
problems in geometric computer vision.
Assuming known camera poses, the problem conceptually boils down to
estimating the unknown parameters of an (explicit or implicit) surface
representation, such that photometric discrepancies between different
views of the scene surfaces are minimized,
\eg~\cite{vu2012}.
A number of recent works have coupled the 3D reconstruction to a
semantic segmentation of the scene, and have shown that, 
unsurprisingly, superior results can be obtained by jointly
optimizing over both geometry and
semantics~\cite{haene2013,savinov2015,ikehata2015}.
A major advantage of such a \emph{semantic 3D reconstruction} approach
is its ability to apply class-specific priors.
Any dense 3D reconstruction algorithm includes a-priori assumptions
to regularize the scene geometry -- in the simplest case some form of
preference for smooth surfaces.
Recovering a semantic segmentation together with the scene geometry
makes it possible to use regularizers that are a lot more expressive,
and take into account the specific geometric properties of different
semantic classes.
For example, one might want to selectively enforce higher smoothness
on roads, where only little texture is available, or straight
boundaries where building walls meet the ground.

In recent work, integrated models have been developed for semantic 3D
reconstruction, but these employ a voxel-based representation of the
scene which limits their resolution.
Increasing the level of detail requires finer volume discretisation,
which in turn increases memory consumption and computational cost,
even with adaptive data structures~\cite{kundu2014,blaha2016}.
Moreover, the subsequent conversion of volumetric occupancy models to
explicit surface meshes typically leads to aliasing artifacts on
surfaces not aligned with the 3D coordinate system.
In this work we present a scalable framework for the refinement of
semantically annotated 3D surface meshes, which starts from a coarse
3D reconstruction and mitigates the mentioned limitations.

State-of-the-art techniques for reconstructing high-quality surface
meshes with fine details employ a two-stage approach
\cite{pons2007,furukawa2008,vu2012}. 
First, a coarse 3D model is generated, usually either with a
volumetric approach followed by marching-cube type mesh extraction, or
by triangulating the raw multi-view point cloud.
The subsequent refinement improves that initial mesh by minimizing the
photometric error \wrt the oriented images.
However, existing refinement procedures are oblivious to scene
semantics. The same regularization is imposed everywhere, without
considering the semantic class or the presence of class transitions.
Further, the result of refinement might require local semantic label changes to maintain semantic consistency.

\myparagraph{Contributions.}
Here we present the first \emph{semantically informed surface refinement}
method for semantic 3D reconstruction.
Like existing methods, ours maximizes photo-consistency to recover
fine details. But it additionally exploits semantic information: (1)
it also maximizes the consistency of labels across semantic
segmentations of the input images; and (2) it constrains the
reconstruction with shape priors that depend on the local semantic
label of the surface.
Our method alternates between variational optimization of the 3D
surface shape and its semantic labeling using Markov Random Field inference.
Consequently, our approach enables the joint refinement of two very different but mutually dependent entities for which information about one entity helps to improve the other.  
We show on a variety of quantitative and qualitative experiments that our approach outperforms existing methods in terms of geometric accuracy as well as semantic label accuracy.

To the best of our knowledge, we present the first mesh refinement method for 3D reconstruction which considers and jointly optimizes semantic label information in order to obtain high-resolution semantic meshes (see \Fig~\ref{fig:front}).
\section{Related Work}\label{sec:related}

Remarkable progress has been made in dense geometry reconstruction
from images. Highly accurate 3D models can now be extracted
automatically using only image data, as witnessed by the results of
multiple influential benchmarks \cite{seitz2006, strecha2008,
  cavegn2014}.

Although these platforms provide an extensive list of well-established
techniques, methods which aim for \emph{semantic 3D reconstruction}
are often not present in their line-up. A possible explanation for
this absence is the missing semantic ground truth. Generic
benchmarking of semantic 3D models is not straightforward, as the
choice of classes depends more directly on the underlying
application. As an example, consider a residential area: if the goal
is to check accessibility, the classification of roads is imperative,
while vegetation may be less important; conversely, for urban climate
or recreation, the vegetation is crucial. Despite this difficulty of
finding a common target output, open semantic 3D datasets have
appeared \cite{sem3D, nyudata}, but so far they do not go beyond point
clouds.

The topic of the present paper is at the interface of \emph{mesh
  refinement}, \emph{mesh-based semantic segmentation} and \textit{semantic 3D reconstruction}. We thus review the relevant literature for these topics.

\myparagraph{Multiview Mesh Refinement.}  Given an initial geometry,
the common approach of variational (multi-view) mesh refinement
formalizes the disagreement between the mesh and the image data in an
energy function and minimizes that energy with gradient descent. Many
works derive the energy in a continuous mathematical framework, \eg
\cite{yezzi2003, gargallo2007}, which in principle allows one to
exactly compute the gradient of the reprojection error and to properly
account for visibility.
Because of the discrete nature of meshes, the gradient then has to be
discretized -- an error-prone process. To circumvent this issue,
\cite{delaunoy2008} directly used the non-smooth surface for the
optimization of the reprojection error. Further methods of this family
are \cite{vu2012, li2015}. Another line of
work bases the refinement on patches instead of surfaces
\cite{furukawa2008, heise2015}.

To align well with the data, the energy function must measure
photo-consistency as a function of the reprojected geometry
\cite{pons2007, vu2012, tylecek2010}. Additionally, further visual cues
can be leveraged, \eg Lambertian surface reflectance \cite{soatto2003,
  delaunoy2008} or contours \cite{tylecek2010}.   
While these method extensions can potentially yield better 3D
models, their complexity increases, with diminishing
returns. In this context, \cite{li2016} propose an efficient mesh
refinement method that is able to determine which model-parts
contribute most towards geometric fidelity, and improve only
those. Finally, current research aims to simultaneously improve also
the camera poses \cite{delaunoy2014}.

To compensate for poor evidence, such as texture-less or noisy regions,
mesh fitting uses adequate regularizers. In the simplest case,
they isotropically penalize strong bending of the surface,
based on its principal curvatures \cite{vu2012}.
The weight of the regularizer can be adapted to the
distinctiveness of the photo-consistency metric \cite{vu2012}.
More context-aware priors also take into account local surface noise
and shape parameters within the denoising process~\cite{li2015}. 

\myparagraph{Semantic Segmentation of Surface Meshes.}
State-of-the-art methods for assigning semantic labels to surface
meshes are based on Conditional Random Fields (CRFs). \cite{Verdie15}
compute geometric features of the mesh in an unsupervised
fashion. Per-face features form the unary potential for labeling the
mesh faces with CRF inference. More related to our approach,
\cite{Rouhan17, valentin13, Kalogerakis10} train classifiers for the
per-face class-conditionals, using texture and/or geometric
features.
All those works combine the per-face unary potential with a generic,
pairwise smoothness term.
On the contrary, we additionally employ geometric surface information
within the labeling process.
Related work on CRF-based mesh labeling in the context of mesh
texturing was presented in \cite{Waechter2014,Lempitsky07}.

\myparagraph{Semantic 3D Reconstruction.}
The goal to jointly infer 3D shape and object classes in a principled
manner, by fitting a coupled model, was initially tackled in
\cite{ladicky2010}. That work was still formulated in 2.5D using
depth maps. True 3D methods appeared only recently, mostly with
volumetric representations \cite{haene2013, bao2013, savinov2015, savinov2016}.
The methods were extended to handle city-scale models \cite{kundu2014,
  blaha2016, vineet2015}; large label sets, also in urban scenes
\cite{cherabier2016}; and thin semantic layers like hair on human
heads \cite{maninchedda2016}.
Departing from the volumetric approach, \cite{cabezas2015} do semantic
3D reconstruction with a (low-resolution) triangle mesh.
The common ground of all these works is that they allow local shape to
influence the appearance-based class labels and vice versa, via
class-specific regularization.

\myparagraph{Relation to our Work.}
To the best of our knowledge, none of the existing \textit{mesh
  refinement} approaches specifically utilizes semantic labels to
impose class-specific shape knowledge. And vice versa, none of the
\textit{semantic 3D reconstruction} techniques applies mesh refinement
to obtain a model with the amount of surface detail present in
high-resolution 3D models. This gap is the starting point for our
work.

\section{Method}\label{sec:semantic_method}

We assume as given a set of calibrated cameras, for which we have both
the intensity images and the semantic segmentations, in the form of
pixel-wise likelihoods for all possible classes.
Furthermore, we assume there is an initial surface mesh, \eg, from
structure-from motion or coarse semantic 3D reconstruction.
The mesh also has per-face semantic labels -- if this is not the case,
they can easily be generated by projecting the per-image class scores
onto the surface and aggregating them with some consensus mechanism.
Our goal is to move the vertices of the initial mesh, and to change
the semantic labels of its faces, until the consistency between the
images is maximized \wrt both photometry and semantic segmentation.
Additionally, we define a set of priors that link geometric shape to
semantic class, and constrain the refinement.

To obtain a tractable algorithm, we split the optimization into two
subproblems, which we then solve independently in an alternating manner.
1) One optimization updates the geometry while keeping the labels fixed. In
that step, the (fixed) labels induce class-specific priors in the
shape, such as for example that building walls tend to be smoother than vegetation.
2) The other optimization relabels the mesh faces, while keeping the
geometry fixed. In that step the surface shape serves as prior that
influences the labeling, \eg vertical faces prefer to be labeled as
building walls.

For the geometric update, we employ a variational mesh refinement
scheme \cite{pons2007,vu2012}, which we extend to include semantic
labels.

The class-specific priors in volumetric reconstruction schemes seek to constrain surface orientation \cite{haene2013,savinov2015,blaha2016,maninchedda2016}. This 
might interfere with the goal of retrieving high detailed surface geometry. In contrast, we leverage the surface curvature and wish to make
the strength of the smoothness prior dependent on the semantic 
class (e.g., high for road, low for vegetation). Further, we want to favor certain edge orientations for faces along class transitions.

For the semantic relabeling, we work on the graph implied by the surface mesh and rely on standard CRF inference.
As feedback from the surface geometry, we include a term that depends
on the face's normal vector, so as to favor labels that are
consistent with the surface orientation.
Empirically the alternation quickly converges to a stable state of labeling and the geometry.
	\subsection{Variational Surface Refinement}\label{sec:prob}

The surface $\nSurf$ is parametrized as a labeled triangle mesh,
represented by the tuple $(\nVertices, \nFaces, \nFaces^\nLabel)$
of vertices $\nVertices$, faces $\nFaces$, and per-face semantic
labels $\nFaces^\nLabel$.
Like most other refinement algorithms we assume that (1) the topology
of the initial mesh does not change during refinement, and (2) the
mesh lies close enough to the true surface to employ local,
gradient-based optimization. The surface is observed by $n$ cameras
with known projections $\nProj_i: \nR^3 \rightarrow \nR^2$, and
associated images $\nImg_i: \nImgDom_i \subset \nR^2 \rightarrow
\nR^d$ with $d\in \{1,3\}$ color channels.  Furthermore, all input
images have been segmented with a semantic per-pixel classifier into a
set of $\nLabelSet=\{1, \ldots,\nNumLabels\}$ different labels.  The
corresponding likelihood images for each label $\nLabel$ are denoted as
$\nLabelImg_i: \nImgDom_i \rightarrow [0,1]$.

Similar to \cite{vu2012}, we compute the refined surface $\nSurf$ as the minimizer
of a variational energy, made up of a data and smoothness terms:
\begin{equation}\label{eq:energy}
  E(\nSurf) = 
    \underbrace{ E_{\text{photo}} + \lambda_1 E_{\text{sem}} }_{\text{data consistency}} +
    \underbrace{ \lambda_2 E_{\text{intra}} + \lambda_3 E_{\text{inter}} }_{\text{smoothness}}
    \enspace.
\end{equation} 
The weights $\lambda_1,\lambda_2,\lambda_3 \in \nRplus$ define the
relative impact of each individual term. The variation of this energy
corresponds to a vector field along the surface, which is used to
iteratively deform the mesh via gradient descent, until convergence.
The data term is divided into a photo-consistency and a semantic
consistency term.  Likewise, the smoothness terms incorporate priors
for semantic intra-class and inter-class dependencies.  In the
following we detail each term individually.
	\subsection{Data Consistency}\label{sec:prob}

\begin{figure}
  \centering
  \includegraphics[width=0.9\columnwidth]{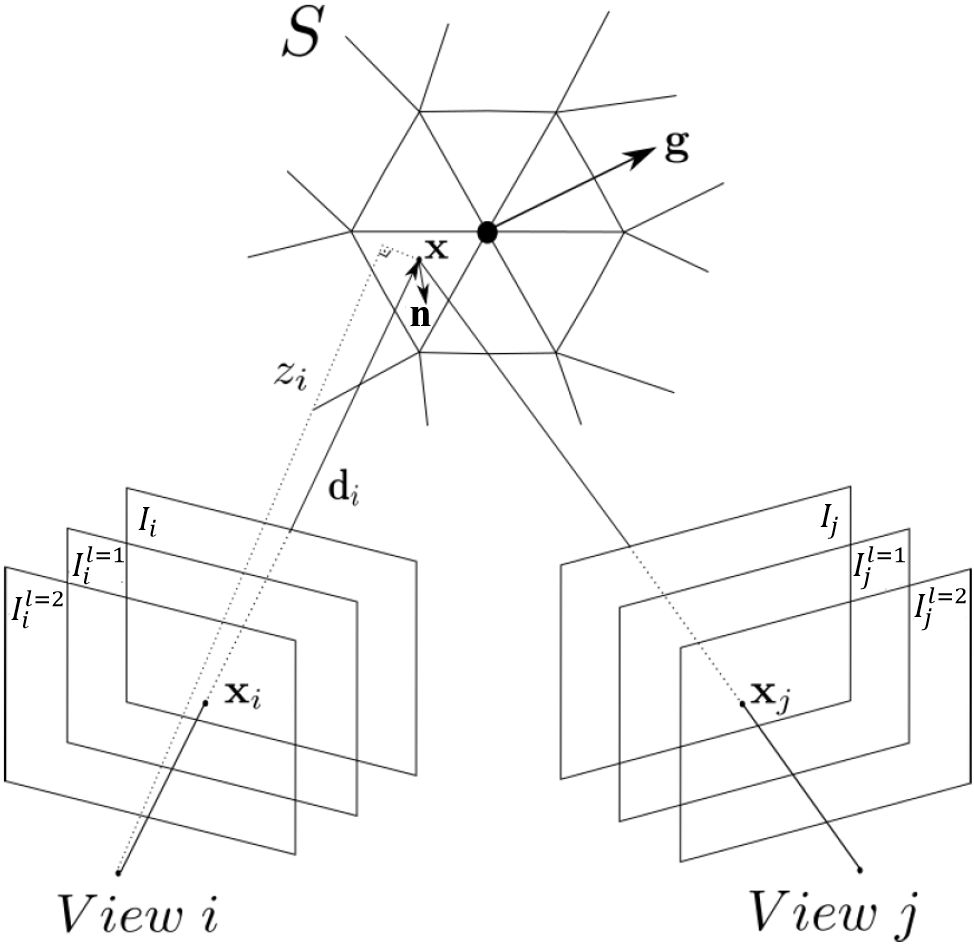}
  \caption{Illustration of the reprojection.}
  \label{fig:concept2}
\end{figure}

We enforce consistency with the input data by two separate energy
terms. One promotes photometric consistency with the images, the other
semantic consistency with the 2D label maps.

\myparagraph{Photometric Consistency.}  The photo-consistency term
minimizes the photometric reprojection error between pairs of camera
images.  Similar to \cite{vu2012} we defined it as
\begin{equation}\label{eq:energy_photocons}
  E_{\text{photo}}(\nSurf)  = \sum_{i,j}\int_{\nImgDom_{ij}} h\big(\nImg_i, \nImg_{ij}\big)(x_i) \;dx_i \enspace,
\end{equation}
where the function $h(I_i, I_{ij})(x_i)$ measures the
photo-consistency between image $I_i$ and $I_j$ at pixel $x_i$, and
$\nImg_{ij} = \nImg_j \circ \nProj_j \circ \nProj_i^{-1}$ is the
reprojection of image $\nImg_j$ into image $\nImg_i$ via the surface
$\nSurf$ and is depicted in \Fig~\ref{fig:concept2}. 
The corresponding image domain $\nImgDom_{ij} \subset
\nImgDom_i$ is induced by the reprojection of image $\nImg_j$.
The energy gradient is given by
\begin{align}\label{eq:grad_photocons}
  \frac{dE_{\text{photo}}(S)}{dX}  
    &= \sum_{i,j}\int_{\nImgDom_{ij}}\phi(x)f_{ij}(x_i)/(\nNormal^T\nDepth_i)\nNormal \;dx_i \\
  f_{ij}(x_i) 
    &= \partial_2 M(x_i)D\nImg_j(x_j)D\nProj_j(x)\nDepth_i \enspace,
\end{align}
in which function $\phi(x)$ weighs the pixels in the back-projected
triangles that contain vertex $x$ according to the barycentric
triangle coordinates. Furthermore, $\nNormal$ is the surface normal, $\nDepth_i$ the
distance from the camera center to the point on the surface, symbol
$D$ denotes the Jacobian of a function, and $\partial_2M$ is the
derivative of the similarity measure \wrt its second
argument. For the similarity measure $h(\cdot)$, we use zero-mean normalized
cross-correlation.
In practice this term enhances the reconstruction of fine details, but
also introduces geometric noise without additional regularization~\cite{vu2012}. 

\myparagraph{Semantic Consistency.}
In the same spirit, the semantic consistency term minimizes the
discrepancies between pairs of 2D semantic segmentation maps
corresponding to different input views.
While accounting for all cameras pairs $i,j$ and all labels
$\nLabel\in\nLabelSet$, the following term measures pixel-wise differences
of class-likelihoods between the two semantic segmentation maps
$\nLabelImg_i$ and $\nLabelImg_j$

\begin{equation}\label{eq:energy_semcons}
  E_{\text{sem}}(\nSurf) = \sum_{\nLabel \in \nLabelSet}\sum_{i,j}\int_{\nImgDom_{ij}} 
    \frac{1}{2} \big(\nLabelImg_i(x_i)-\nLabelImg_{ij}(x_i)\big)^2 \; dx_i \enspace.
\end{equation}
This term and its derivative is identical to \Eq~\eqref{eq:grad_photocons} with the difference that the similarity measure $h(\cdot)$ is the sum of squared differences and the comparison is done between label likelihoods instead of color values.
	\subsection{Smoothness of Geometry}\label{sec:prob}
 
Smoothness of the refined surfaces is encouraged by two
terms, one for intra-class, and one for inter-class regularity.
 
\myparagraph{Intra-class Smoothness.}
Here we use the classical thin-plate minimal curvature regularization,
generalized for the multi-class setting.
In this way, we can enforce different levels of smoothness for different
classes. For instance, facades are in general smoother than
vegetation.
Furthermore, class transitions mostly coincide with high surface
curvature (\eg from ground to building).
Consequently, we introduce a smoothing weight for each class and our
intra-class smoothness term reads as
\begin{equation}\label{eq:energy_intra_smooth}
  E_{\text{intra}}(\nSurf) = 
    \sum_{\nLabel\in\nLabelSet}\sum_{\nVertex \in \nVertices_\nLabel} \omega_\nLabel
    \frac{1}{2} \Big(\kappa_1(\nVertex) + \kappa_2(\nVertex) \Big) \enspace,
\end{equation}
where $\nVertices_\nLabel \subset \nVertices$ corresponds to vertices that
encompass a semantically homogeneous one-ring-neighborhood with label
$\nLabel$, and $\omega_\nLabel$ a class-specific weight factor.
Note that this regularizer can easily be extended to include
image-based information, e.g. adapt the amount of smoothing along edges
in the input image, as they often align with object edges (as for
example done in \cite{Wu2011}).

\begin{figure} 
  \centering
    \includegraphics[width=0.49\columnwidth]{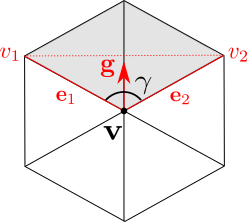}
  \caption{ Top view on a two-label one-ring-neighborhood of
    $\mathbf{v}$. Visualization of the construction of the gradient
    vector $\mathbf{g}$ used for forcing smooth transition
    boundaries.}
    \label{fig:smoothness2}
\end{figure}

\myparagraph{Inter-class Smoothness.} 
The inter-class smoothness accounts for regularity along class
boundaries. On a discrete mesh, these boundaries are represented by
sequences of adjacent edges, each adjoining two faces with different
labels. To enforce smoothness of these boundaries, we penalize
angular deviations along class transitions by minimizing the
energy
\begin{equation}\label{eq:energy_inter_smooth2}
  E_{\text{inter}}(\nSurf) = 
    \sum_{\nVertex \in \nVertices_T}(\pi - \gamma(\nVertex))^2 \enspace,
\end{equation} 
in which $\nVertices_T \subset \nVertices$ is the set of vertices
featuring a two-label one-ring-neighborhood, where faces with equal
labels are direct neighbors. Class
transitions in such a triangle fan are defined by two edges
$\nEdge_1$, $\nEdge_2$ and the corresponding vertices
$\nVertex$, $\nVertex_{1}$, $\nVertex_{2}$, see \Fig~\ref{fig:smoothness2}. 
The energy of \Eq~\eqref{eq:energy_inter_smooth2} is minimal if the angle
$\gamma(\nVertex)$ between $\nEdge_1$ and $\nEdge_2$ is
$\pi$. Note that in the current definition we do not consider triangle
fans with more than two labels because a separate modeling of these rare cases is not beneficial. 

\myparagraph{Discussion.}
When all semantically related terms and relations are neglected, \ie we consider only one label, the intra-class smoothness term in \Eq~\eqref{eq:energy_intra_smooth} simplifies
to
\begin{equation}\label{eq:semprior}
  E_{\text{smooth}}(\nSurf) = \sum_{\nVertex \in \nVertices} \frac{1}{2}
    \Big(\kappa_1(\nVertex) + \kappa_2(\nVertex) \Big)\enspace,
\end{equation} 
with $\kappa_1(\nVertex)$ and $\kappa_2(\nVertex)$ being the principal
curvatures at vertex $\nVertex$.  Together with the remaining
photo-consistency term in \Eq~\eqref{eq:energy_photocons} the overall energy
reduces to
\begin{equation}\label{eq:energy_baseline}
  E_{\text{baseline}}(\nSurf) = E_{\text{photo}}(\nSurf) + \lambda E_{\text{smooth}}(\nSurf) \enspace.
\end{equation} 
The minimization of this simplified energy corresponds to the purely
geometric refinement method described in \cite{vu2012}, which we use
as baseline for comparisons.  Our geometry refinement can be seen as
generalization of \cite{vu2012} that incorporates semantic labels
and a corresponding set of rich, semantic priors which allow to favor different class-dependent surface properties.

\myparagraph{Minimization.} The sum of the derivatives of all energy terms defines for each vertex a 3D direction in which it should be moved to improve the energy.
Together with a fixed step width this defines a single gradient descent step which we iterate until convergence in alternation with the semantic relabeling.
	\subsection{Semantic Relabeling}

The variational surface refinement changes the mesh geometry.
As individual faces move, their labels may become inconsistent with
those given by the 2D semantic likelihoods of the input images.
In order to minimize such inconsistencies we relabel the faces after a
fixed number of geometric refinement iterations.
The relabeling is formulated as an energy minimization in a Markov Random
Field.
Each face corresponds to a node in the MRF graph, sharing three node interactions
with its adjacent faces.
Let $\nFaces$ be the set of faces and $\nLabelSet$ the set of
potential class labels. The goal is to derive a labeling
$\nLabeling\in\nFaces^\nLabel$ that assigns a label $\nLabel_\nFace \in \nLabelSet$ to each face $\nFace\in\nFaces$, such that the energy $E(\nLabeling)$ is minimized.
More precisely, our energy has the form
\begin{align}
  E(\nLabeling) &= \sum_{\nFace\in\nFaces} E_{\text{data}}(\nLabel_\nFace) 
    + \mu_1 \sum_{\nFace\in\nFaces} E_{\text{geo}}(\nNormal_\nFace,\nLabel_\nFace) \nonumber \\ 
  & + \mu_2 \sum_{\nFace\in\nFaces, g\in\nNeigh_\nFace} 
      E_{\text{smooth}}(\nLabel_\nFace, \nLabel_g) \enspace,
\end{align}
where the weights $\mu_1, \mu_2$ control the contribution
of the label-dependent geometry prior $E_{\text{geo}}$ and the label
smoothness prior $E_{\text{smooth}}$.
The data term
\begin{equation}
  E_{\text{data}}(\nLabel_\nFace) = 
    -\log \left(\sum_{i} \int_{\Psi_i} \nImg^{\nLabel_\nFace}_i(x_i) \; dx_i \right)
\end{equation}
integrates the likelihoods of class $\nLabel_\nFace$ in the likelihood images
$\nImg_i^{\nLabel_\nFace}$.
Per image, integration is carried out over the domain $\Psi_i$,
defined by the area of the reprojected face.
Note that by integration in image space we exploit the same benefits
as during geometric refinement: we put emphasis on large faces and
images close to the surface, while reducing the influence of
observations from slanted viewing angles.

Analogous to the use of semantically modulated priors during geometric
refinement, we now employ a geometry-dependent prior for the mesh
labeling.
In this way we retain a tight coupling between the semantic and geometric
optimization steps.
The prior is dependent on the face normal and penalizes class labels
which contradict the surface orientation.
For example, labeling a face with vertical normal as a facade would
induce a higher cost than assigning it the class ground.
Due to the wide range of possible normals of our classes (ground, facade, roof, vegetation), we found it best to define individual penalties as conservatively parametrized step functions: 
\begin{equation}
  E_{\text{geo}}(\nNormal_\nFace,\nLabel_\nFace) =
  \begin{cases}    
    A_\nFace \quad \text{if} \quad \|\measuredangle(\nNormal_\nFace, \nNormal_z) - \beta(\nLabel_\nFace) \| > \alpha(\nLabel_\nFace)  \\
    0 \quad \text{otherwise}
  \end{cases} \enspace.
\end{equation}   
Here $\nNormal_z$ corresponds to the gravity vector
(typically $(0,0,1)^{T}$), $\nNormal_\nFace$ is the face normal and $A_\nFace$
is the triangle size, to account for the surface area.
The parameter $\beta$ specifies the ideal (set of) class-wise
normal(s) expressed with respect to $\nNormal_z$.
The parameter $\alpha$ determines the range of angles for which
penalties are imposed.
The parameters used in our experiments are given in Table~\ref{tab:paramslabeling}.

Given a face $\nFace$, and another face $g$ in its one-ring neighborhood
$\nNeigh_\nFace$, the pair-wise smoothness term is defined as
\begin{equation}
  E_{\text{smooth}}(\nLabel_\nFace,\nLabel_g)=
  \begin{cases}     
    A_f \quad \text{if} \quad \nLabel_g \neq \nLabel_\nFace \\    
    0 \quad \text{otherwise} 
  \end{cases}.
\end{equation}   
As before, the triangle area $A_\nFace$ serves as weight to increase
contributions of larger triangles.

To find a (local) minimum of $E(\nLabeling)$ we run loopy belief
propagation~\cite{frey}.
In all experiments we set $\mu_1=0.35$, $\mu_2=0.5$.
\begin{table}
  \centering
  \renewcommand{\arraystretch}{1.1}
  \begin{tabular}{ c | c | c | c | c }
    & \textbf{roof} & \textbf{vegetation} & \textbf{ground} & \textbf{facade} \\
    \hline
    \hline
    $\alpha(\nLabel_\nFace) [^\circ]$ & 60 & 180 & 30 & 30 \\
    \hline
    $\beta(\nLabel_\nFace)  [^\circ]$ & 0  & 0   & 0  & 90 
  \end{tabular}
  \vspace{0.2cm}
  \caption{Parameters of class-dependent geometric prior.}
  \label{tab:paramslabeling}
\end{table}
\section{Experiments}\label{sec:experiments}

All experiments were performed on a machine with 64 GB of RAM and a
12-core \textit{Intel Xeon E5} CPU at 2.7~GHz. We start with a
quantitative evaluation on a synthetic scene, and then go on to
reconstruct four challenging real world scenarios featuring different sensors and camera network configurations.

\begin{figure}
  \centering
  \includegraphics[clip=true, trim=20 115 20 135,width=0.99\columnwidth]{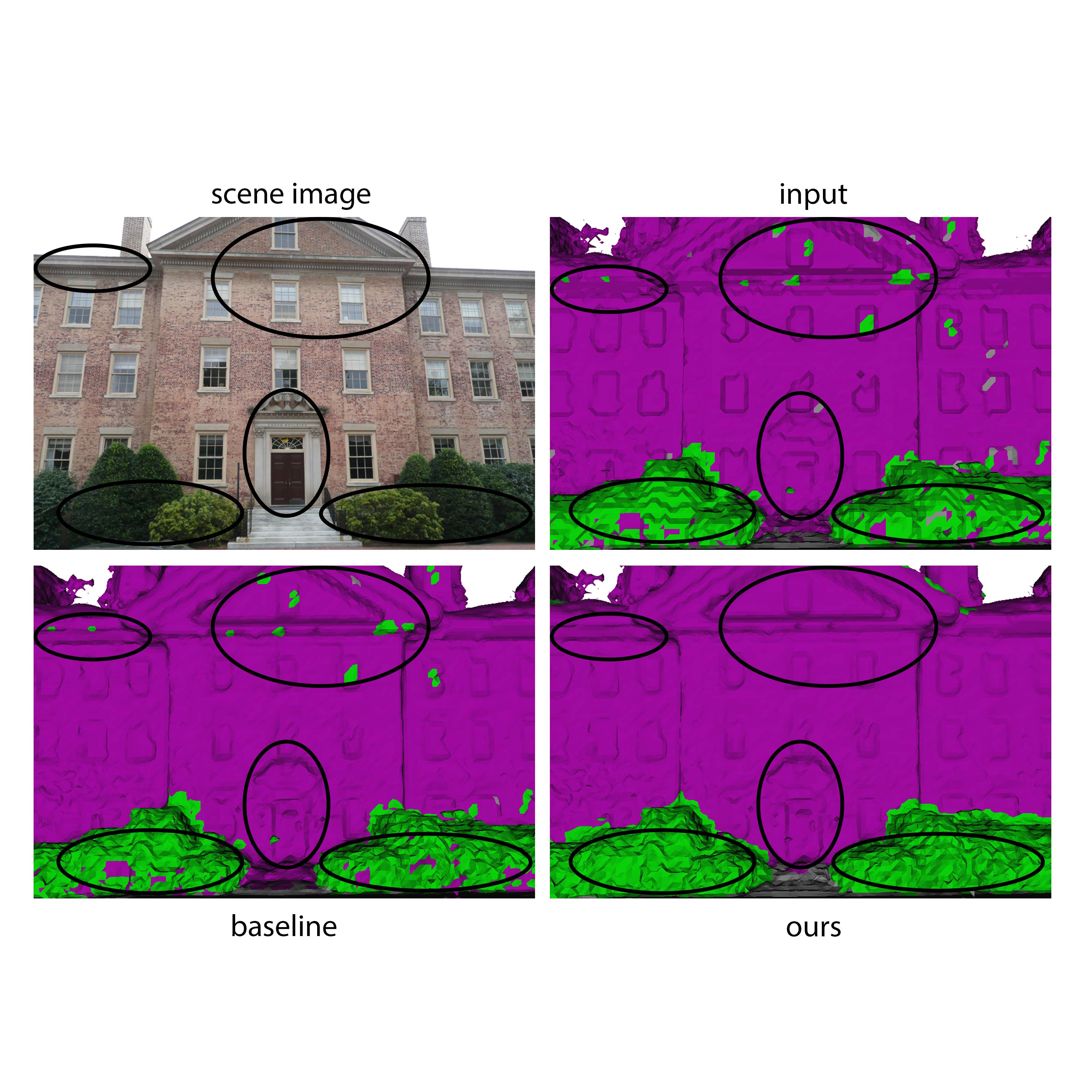}
  \vspace{0.1cm}
  \caption{Results of the terrestrial \textit{Southbuilding} data
    set. The proposed method outperforms our input \cite{blaha2016}
    and baseline \cite{vu2012} model. Exemplary improvements are
    highlighted with black circles.}
\label{fig:south}
\end{figure}

\myparagraph{Data.} The processed datasets comprise vertical and
oblique aerial views of urban areas, as well as a terrestrial outdoor scenario. For a quantitative verification
and a comparison to state-of-the-art methods, we process
\textit{SynthCity3} from \cite{cabezas2015}, for which a labeled
ground truth model is available. To test our algorithm on real world
data, we further process three image sets covering Enschede
(Netherlands)\cite{slagboom}, Dortmund (Germany)\cite{isprs} and \textit{Southbuilding} (Fig.~\ref{fig:south}).
For the very large \textit{SynthCity3}
and Enschede datasets we process two sub-patches (in the following referred to as \textit{SynthCity3 A}/\textit{B} and Enschede A/B).
For the real-world scenarios, geometric ground truth is not available. However, to quantitatively check the correctness of our models, we render the surface semantics and compare the results to hand-labeled, semantic ground truth images. \Tab~\ref{tab:input} shows the characteristics of the input data at a glance.

\begin{figure}
  \centering
  \includegraphics[clip=true, trim=0 75 0 0, width=0.32\columnwidth]{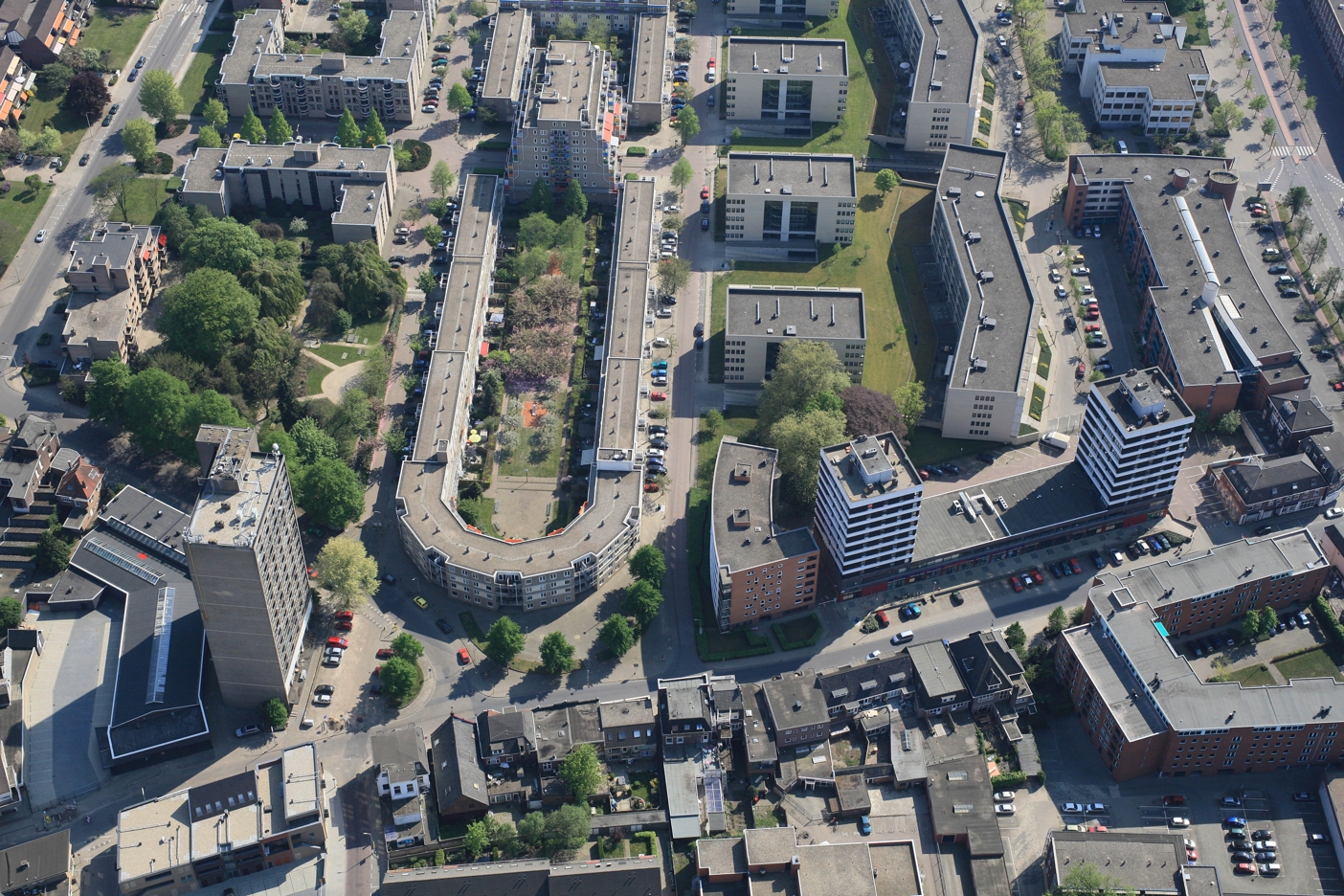}
  \includegraphics[clip=true, trim=0 75 0 0, width=0.32\columnwidth]{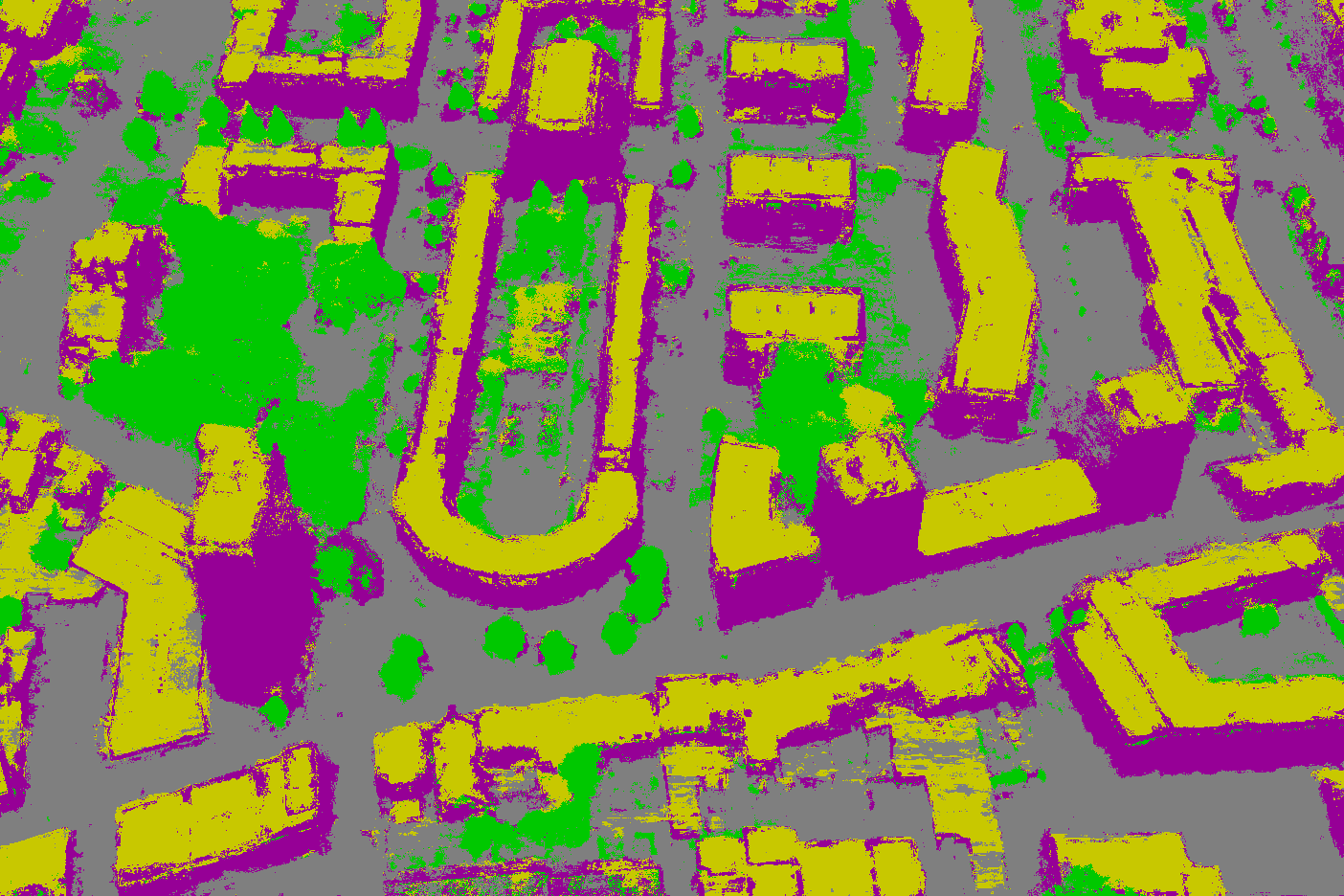}
  \includegraphics[clip=true, trim=50 61 50 10,width=0.34\columnwidth]{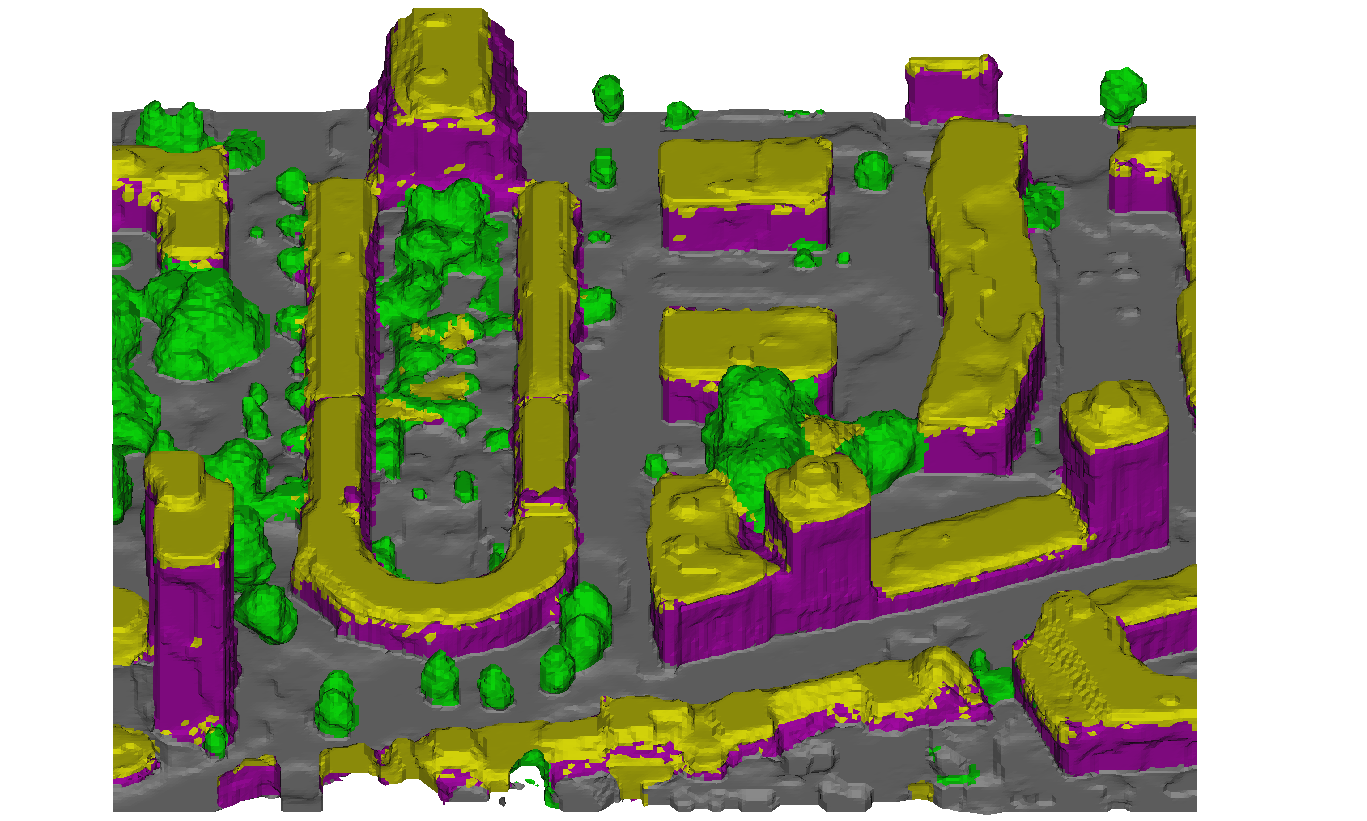}
  \caption{Input data of our algorithm (Enschede A), comprising intensity images
    (grayscale or RGB), semantic segmentation maps and an initial
    semantic 3D model with consistent topology
    (\textit{left-to-right}). The segmentation map is visualized by
    the class with maximum likelihood among: ground (\textit{gray}),
    facade (\textit{purple}), roof (\textit{yellow}), vegetation
    (\textit{green}).}
\label{fig:input_data2}
\end{figure}

Our algorithm requires three types of inputs: (1)~intensity images,
\ie RGB or grayscale, (2)~semantic segmentation maps of those images
with a per-pixel likelihood for each class, and (3)~an initial,
semantically annotated 3D surface with consistent topology. The semantic segmentations (2) are obtained from a MultiBoost classifier \cite{benbouzid2012} trained on a few manually labeled images. For our scenarios, we choose four mutually exclusive
labels: ground, facade, roof and vegetation. The input surface (3)~is
generated using the semantic 3D reconstruction of \cite{blaha2016},
followed by marching-cube mesh conversion. \Fig~\ref{fig:input_data2}
illustrates a sample of our input data from the Enschede data set.

\begin{table}
\vspace{0.1cm}
  \centering
  \small
  \renewcommand{\arraystretch}{1.3}
  \begin{tabularx}{\columnwidth}{ l|c|c|c }
		\textbf{Data set} & \textbf{Resolution [pix]} & \textbf{GSD [m]} & \textbf{\# of images}  \\
		\hline
		\hline
 		\textit{SynthCity3} A & 1416 x 1062 & N/A & 15 \\
 		\hline
 		\textit{SynthCity3} B & 1416 x 1062 & N/A & 15 \\
 		\hline
 		Enschede A & 1404 x 936 & 0.4 & 15 \\
		\hline
 		Enschede B & 1404 x 936 & 0.4  & 14\\
 		\hline
 		Dortmund & 1363 x 1020 & 0.60 & 12\\
 		\hline
 		\textit{Southbuilding} & 768 x 576 & N/A & 15\\
\end{tabularx}
\vspace{0.1cm}
\caption{Technical specifications of the processed image sets. The GSD corresponds to the average pixel footprint of the vertical and oblique aerial images. In \textit{SynthCity A} and \textit{B}, this value has no metric unit. In the terrestrial \textit{Southbuilding} images, it varies greatly across the scene.}
\label{tab:input}
\end{table}

\myparagraph{Quantitative Evaluation.}  We evaluate our method in
terms of geometric and semantic accuracy.  As baseline, we use our own
reimplementation of the state-of-the-art algorithm for mesh refinement
\cite{vu2012}.  Further, the proposed approach can generally be
considered as a post-processing step for semantic 3D modeling
algorithms like \cite{haene2013, blaha2016, savinov2015,
  savinov2016}. For this reason, we also compare against our
initialization, the output of \cite{blaha2016}.

In terms of geometry, we are interested in two aspects: how much is
the overall improvement compared to the input model, and how do we
perform against the baseline mesh refinement? The first aspect
confirms the need for refinement after semantic 3D reconstruction, and
at the same time checks whether the input mesh is good enough to serve
as starting value for local optimization. The second comparison
assesses our contribution, answering the question: \textit{Does the 
semantic information improve the mesh?} Moreover, checking the
semantic correctness verifies if the geometry refinement and
input likelihoods can be leveraged for semantic relabeling.

For geometric verification we use \textit{SynthCity3 A} and \textit{B}.
Each of those represents a building block of an urban scenario. For a
fair comparison, we first empirically determine the parameters of the
baseline that lead to the most accurate 3D model. This configuration
is then fixed, and only the parameters of our additional semantic
terms are changed for further fine-tuning. Finally, we run five
iterations for both methods. Per iteration, we perform a geometric
update for the baseline, and a sequential geometric and semantic
update for our method.  Our method achieves the highest geometric
accuracy in this test (\Tab~\ref{tab:performance}). Due to the
synthetic nature of the reference model, the values are relative.
To augment them with an absolute metric unit, we measure and average
the dimensions of comparable real world city blocks, and scale our
models accordingly.

\begin{figure*}
  \centering
  \includegraphics[clip=true, trim=180 550 350 200, width=0.99\textwidth]{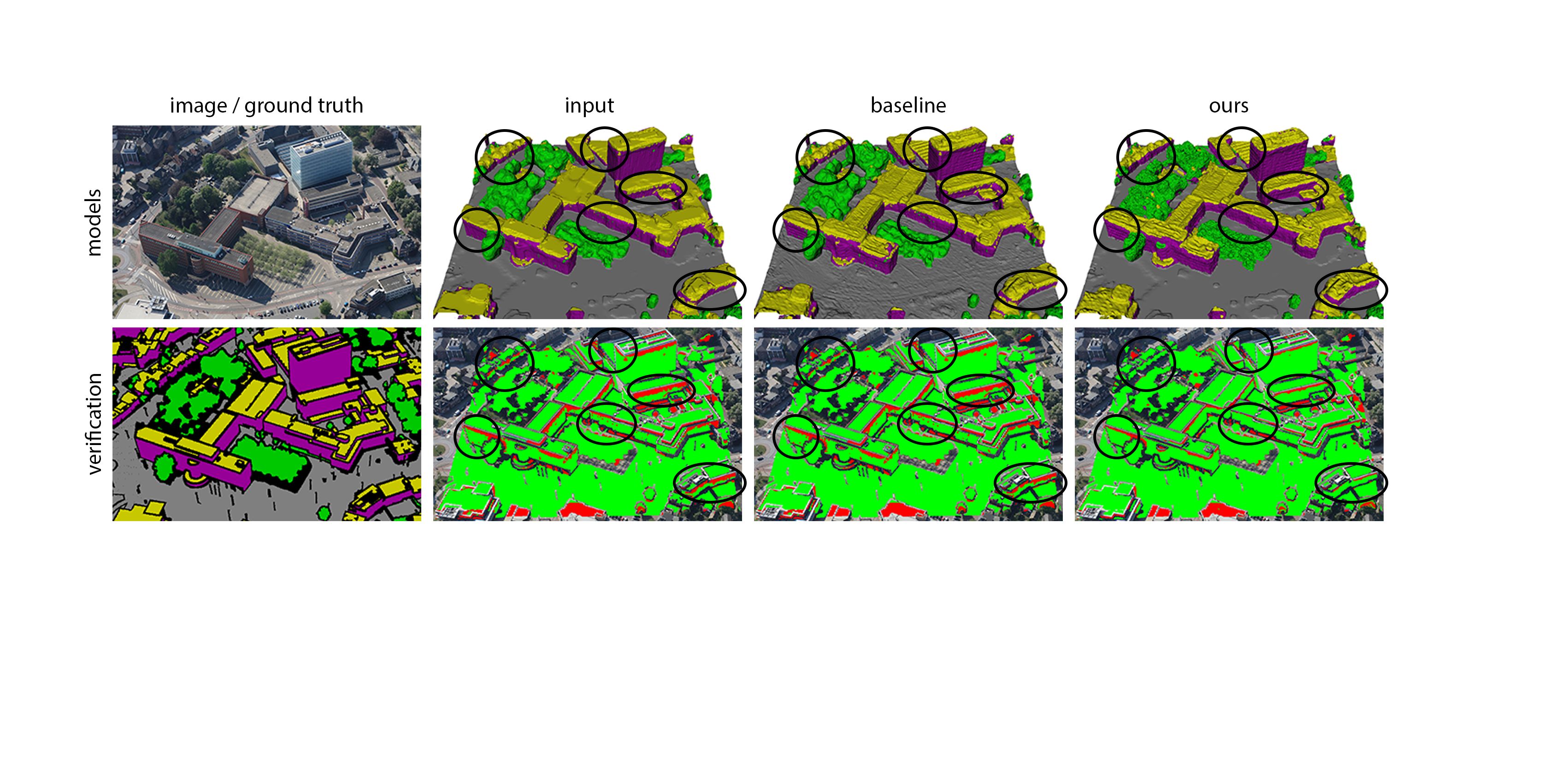}
  \caption{Quantitative evaluation of the semantic correctness
    (Enschede B). \textit{Top}: scene image, input model
    \cite{blaha2016}, baseline model \cite{vu2012}, ours
    (\textit{left-to-right}).
    \textit{Bottom}: ground truth, error plot input,
    error plot baseline, error plot ours. Misclassified pixels are
    shown in red. Exemplary improvements are highlighted with 
    black circles.}
\label{fig:ver}
\end{figure*}

In contrast to geometry, the semantic correctness of real-world models
was quantitatively checked for all scenes. To obtain ground truth
data, we select a representative image in each scenario and manually
label it. The semantic 3D reconstructions are then projected, and
compared to the ground truth in terms of \textit{average accuracy} and
\textit{overall accuracy}\footnote{The \textit{overall accuracy}
  corresponds formally to $A_O = \frac{\sum_{i}c_{ii}}{N}$, where
  $c_{ii}$ are the entries of the confusion matrix and $N$ is the
  number of pixels. The \textit{average accuracy} corresponds to the
  average of the user's accuracy, which in turn is defined as $A_U =
  \frac{c_{ii}}{N_i}$, \ie the ratio between the correct
  classified pixels of a certain class ($c_{ii}$) and the overall
  number of pixels of the same class ($N_i$).}. This procedure is
illustrated for the Enschede B data set in
\Fig~\ref{fig:ver}. \Tab~\ref{tab:performance} summarizes the numeric
results. Again, we outperform our input and baseline method in all
data sets.

\begin{table*}
\setlength\tabcolsep{0.22cm}
\centering
\small

\begin{tabular}{ l|l|l|c|c|c }
\textbf{Data set} & \textbf{Modality} & \textbf{Performance Measure} & \textbf{\cite{blaha2016}} & \textbf{\cite{vu2012}} & \textbf{Ours} \\ \hline \hline

\multirow{4}{*}{\textit{SynthCity3 A}}
& \multirow{2}{*}{Geometry}

& Mean distance to ground truth [relative] & 0.0076 & 0.0064 & \textbf{0.0055} \\
& & Mean distance to ground truth [m] & 0.52 & 0.44 & \textbf{0.38} \\ \cline{2-6}

& \multirow{2}{*}{Semantics}
& Average accuracy [\%] & 82.6 & 82.8 & \textbf{88.8} \\
& & Overall accuracy [\%] & 85.2 & 85.6 & \textbf{86.1} \\ \hline

\multirow{4}{*}{\textit{SynthCity3 B}}
& \multirow{2}{*}{Geometry}

& Mean distance to ground truth [relative] & 0.0121 & 0.0107 & \textbf{0.0090} \\
& & Mean distance to ground truth [m] & 0.84 & 0.74 & \textbf{0.62} \\ \cline{2-6}

& \multirow{2}{*}{Semantics}
& Average accuracy [\%] & 83.6 & 83.8 & \textbf{90.0} \\
& & Overall accuracy [\%] & 86.2 & 86.5 & \textbf{88.7} \\ \hline

\multirow{2}{*}{Enschede A (Netherlands)}
& \multirow{2}{*}{Semantics}
& Average accuracy [\%] & 78.8 & 78.8 & \textbf{83.3} \\
& & Overall accuracy [\%] & 82.6 & 82.7 & \textbf{85.2} \\ \hline

\multirow{2}{*}{Enschede B (Netherlands)}
& \multirow{2}{*}{Semantics}
& Average accuracy [\%] & 89.5 & 89.7 & \textbf{93.5} \\
& & Overall accuracy [\%] & 90.4 & 90.6 & \textbf{94.1} \\ \hline

\multirow{2}{*}{Dortmund (Germany)}
& \multirow{2}{*}{Semantics}
& Average accuracy [\%] & 86.5 & 86.6 & \textbf{87.6} \\
& & Overall accuracy [\%] & 92.3 & 92.4 & \textbf{92.7} \\ \hline

\multirow{2}{*}{\textit{Southbuilding}}
& \multirow{2}{*}{Semantics}
& Average accuracy [\%] & 81.9 & 78.7 & \textbf{94.5} \\
& & Overall accuracy [\%] & 93.8 & 93.8 & \textbf{98.0} \\ \hline

\end{tabular}

\vspace{0.2cm}
\caption{Quantitative evaluation of our method. Best performance is shown in bold.}

\label{tab:performance}
\end{table*}

\begin{figure*}
  \centering
  \includegraphics[clip=true, trim=120 400 60 150, width=0.99\textwidth]{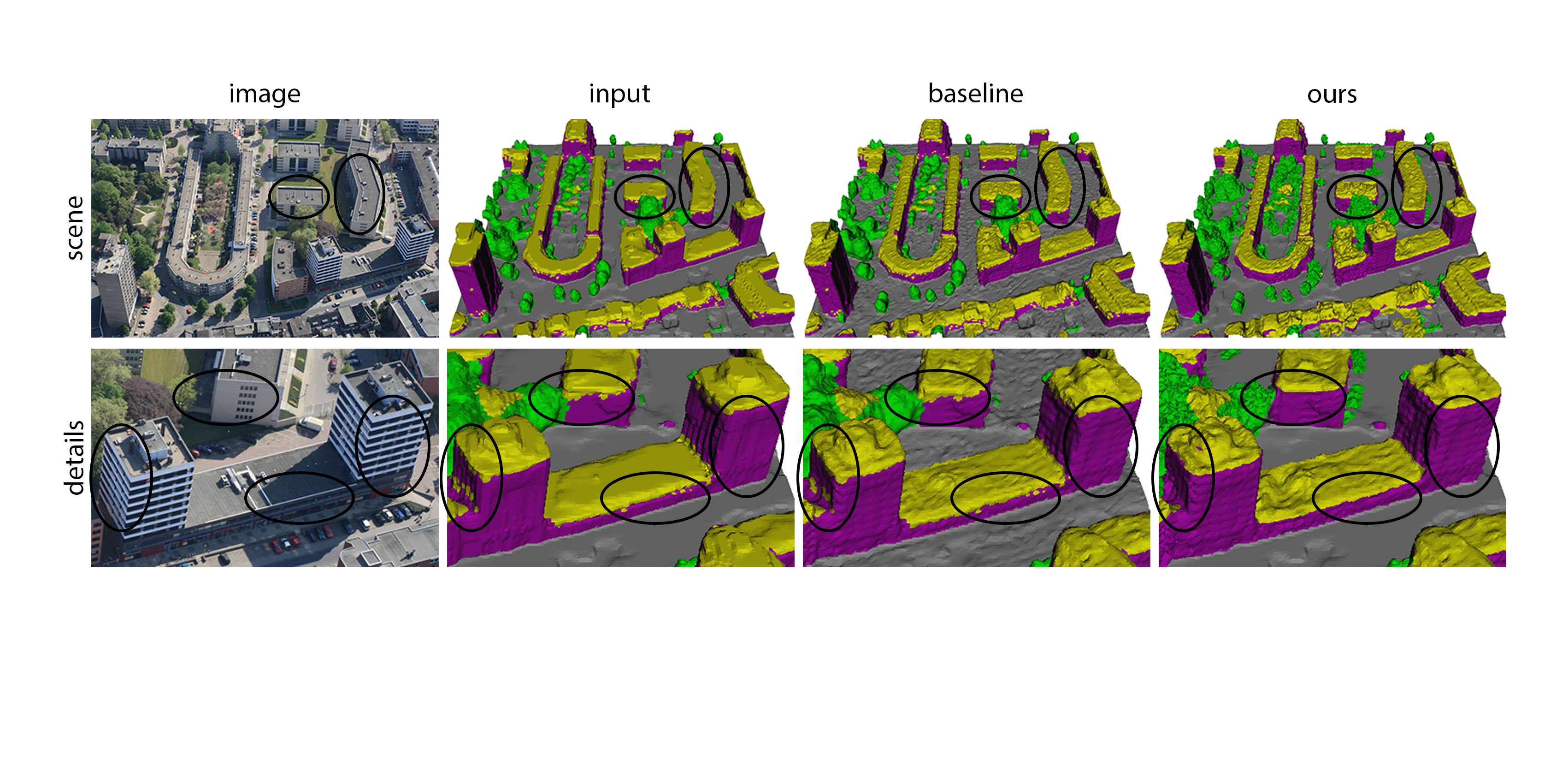}
  \caption{Qualitative evaluation of our method based on models of the
    Enschede A data set (\textit{top}) and corresponding details
    (\textit{bottom}). \textit{Left-to-right}: Input model
    \cite{blaha2016}, baseline model \cite{vu2012}, ours. Notice the
    high scene fidelity, and, at the same time an adaptive,
    class-specific surface regularization, clean class transitions and
    less noisy semantics in our model. Exemplary improvements are
    highlighted with black circles.}
\label{fig:quality}
\end{figure*}

\myparagraph{Qualitative Evaluation.} Jointly exploiting geometry and
semantics during the surface refinement allows for a more steerable
procedure. The additional degrees of freedom can be leveraged to
obtain qualitatively better results, as we will demonstrate on the basis
of an urban test scenario (\Fig~\ref{fig:quality}). The class facade
appears vertical and flat in our input models and suffers from
aliasing, due to the preceding volumetric representation. In contrary,
our method recovers fine structures (\eg windows), removes artifacts
and performs adequate smoothing. For the second class vegetation,
trees appear as blobs in our initial geometry. Due to the highly
undulated surface structure of trees within our method smoothness
regularization is limited to a large degree and we strive for high
data alignment. This leads to more realistic reconstructions of
vegetation areas. Finally, we decide to keep geometry annotated by the
class ground close to the highly regularized input models for the
following reasons: (1) refining the geometry would mostly reconstruct
cars which are dynamic objects and not of interest in terms of the
static scene; (2) the ground does only suffer minor aliasing
artifacts, since the scenes are aligned with the gravity vector.

\myparagraph{Beyond Aerial Reconstruction.} 3D reconstruction of urban
habitates from aerial data is a major application of 3D modeling and
scene interpretation from images today, which was the reason and
motivation to test our method primary on these type of data. However,
to show the versatility of the proposed method we additionally perform
an evaluation on the terrestrial \textit{Southbuilding} dataset,
featuring very different sensor and scene characteristics. As for the
aerial settings, we outperform our input and the
baseline method. \Tab~\ref{tab:performance} summarizes
the quantitative results, \Fig~\ref{fig:south} show visual
differences.
\section{Conclusion}\label{sec:conclusion}

We proposed a method for the joint geometric and semantic
refinement of labeled 3D surfaces meshes. Our algorithm leverages photometric and semantic image information in an integrated manner to refine the geometry of an inital surface mesh. Furthermore, we exploit semantic
information for class-aware regularization within geometric refinement, and
use geometric shape to improve the semantic labeling.
Our optimization scheme alternates between variational surface refinement and MRF
inference for relabeling. In a broad sense, our method corresponds to
a generalization of pure geometric surface refinement, which
incorporates semantic labels and a rich set of corresponding semantic
priors. At the same time it can be seen as a multi-view consistent
semantic segmentation in 3D. Combining both aspects leads to
superior, more detailed and interpreted 3D models.

Our method is not limited to outdoor views of urban scenes, like those
tested within this paper. In future work, we would like to experiment with different
types of data, such as imagery of indoor scenarios. As
the scene characteristics and data quality varies greatly across
different test beds and sensors, we would like to explore methods for
data-driven, automatic balancing of the individual terms in our
framework.

{\small
\bibliographystyle{ieee}
\bibliography{paper}
}

\end{document}